\pdfoutput=1
\documentclass[11pt]{article}

\usepackage[]{acl}

\usepackage{times}
\usepackage{latexsym}

\usepackage[T1]{fontenc}
\usepackage[utf8]{inputenc}

\usepackage{microtype}

\usepackage{inconsolata}

\usepackage{xspace}
\usepackage{amsfonts}

\usepackage{algorithm}
\usepackage[noend]{algpseudocode}
\usepackage{amsmath}
\usepackage{booktabs}
\usepackage{color}
\usepackage{graphicx}

\usepackage{makecell}
\usepackage{multirow}

\usepackage{arydshln}
\usepackage{enumitem}

\setlist[itemize]{leftmargin=*}

\newcommand{\red}[1]{\textcolor{red}{#1}}
\newcommand{\blue}[1]{\textcolor{blue}{#1}}

\definecolor{blue-violet}{rgb}{0.54, 0.17, 0.89}

\def \pos {\mathnormal{positive}}
\def \neg {\mathnormal{negative}}
\def \neu {\mathnormal{neutral}}

\def \entity {\mathnormal{entity}}
\def \aspect {\mathnormal{aspect}}

\def \e {\mathnormal{e}}
\def \a {\mathnormal{a}}
\def \o {\mathnormal{o}}
\def \p {\mathnormal{p}}
\def \s {\mathnormal{s}}

\def \i {\mathnormal{i}}

\def \r {\mathbf{r}}
\def \d {\mathbf{d}}
\def \h {\mathbf{h}}

\def \W {\mathbf{W}}
\def \L {\mathbf{L}}

\def \CQ {\mathcal{Q}}
\def \CE {\mathcal{E}}
\def \CA {\mathcal{A}}
\def \CO {\mathcal{O}}
\def \CP {\mathcal{P}}

\def \CLS {\mbox{CLS}}
\def \SEP {\mbox{SEP}}

\newcommand{\cparagraph}[1]{\vspace{1mm}\noindent\textbf{#1}}

\title{Entity-Aspect-Opinion-Sentiment Quadruple Extraction for Fine-grained Sentiment Analysis}

\author{Dan Ma, Jun Xu, Zongyu Wang, Xuezhi Cao, Yunsen Xian \\
 Meituan \\ 
\texttt{madan07,xujun12,wangzongyu02,caoxuezhi,xianyunsen}@meituan.com}

\begin{document}
\maketitle
\begin{abstract}

Product reviews often contain a large number of implicit aspects and object-attribute co-existence cases. 
Unfortunately, many existing studies in Aspect-Based Sentiment Analysis (ABSA) have overlooked this issue, which can make it difficult to extract opinions comprehensively and fairly. 
In this paper, we propose a new task called Entity-Aspect-Opinion-Sentiment Quadruple Extraction (EASQE), which aims to hierarchically decompose aspect terms into entities and aspects to avoid information loss, non-exclusive annotations, and opinion misunderstandings in ABSA tasks.
To facilitate research in this new task, we have constructed four datasets (Res14-EASQE, Res15-EASQE, Res16-EASQE, and Lap14-EASQE) based on the SemEval Restaurant and Laptop datasets. We have also proposed a novel two-stage sequence-tagging based Trigger-Opinion framework as the baseline for the EASQE task.
Empirical evaluations show that our Trigger-Opinion framework can generate satisfactory EASQE results and can also be applied to other ABSA tasks, significantly outperforming state-of-the-art methods. We have made the four datasets and source code of Trigger-Opinion publicly available at https://github.com/XXX to facilitate further research in this area.

\end{abstract}

% \begin{wraptable}{r}{0.5\textwidth}
 % \centering
 \begin{table}[htbp]
\vspace{-1mm}
 \vspace{-1mm}
  % \captionsetup{skip=5pt} % adjust the spacing here
  \small
  \begin{tabular}{c|l|c|c|c|c}
    \toprule
   \multicolumn{2}{c|}{\bf Datasets} & {\bf \#S$^a$ }& {\bf \#Q$^b$ } & {\bf Co Pct.$^c$}& {\bf N Pct.$^d$} \\
    \midrule
   \multirow{3}{*}{14Res} & Train & 1259 &  2526 & 6.04\% & 11.64\% \\
   % 294
   & Dev & 315  & 594 & 4.13\%  & 8.59\% \\
   % 51
   & Test & 493  & 1102 & 5.48\%  & 11.52\% \\
   % 127
   \hline
   \multirow{3}{*}{14Lap} & Train & 899 & 1492 & 13.13\%  & 13.81\% \\
   % 206
   & Dev & 225 & 408 & 4.00\%  & 16.42\% \\
   % 67
   & Test & 332  & 575 & 14.16\%  & 15.13\% \\
   % 87
   \hline   
   \multirow{3}{*}{15Res} & Train & 603 & 1093 & 3.32\%  & 5.93\% \\
   % 65
   & Dev & 151 & 264 & 4.64\%  & 9.47\% \\
   % 25
   & Test & 325 & 564 & 5.23\%  & 13.83\% \\
   % 78
   \hline   
   \multirow{3}{*}{16Res} & Train & 863  & 1547 & 5.91\% & 7.89\% \\
   % 122
   & Dev & 216  & 385 & 6.02\% & 9.84\% \\
   % 38
   & Test & 328 &  597 & 9.76\% & 13.23\% \\
   % 79
  \bottomrule
  \multicolumn{6}{l}{$^a$: \#S denotes the number of sentences.}\\
  \multicolumn{6}{l}{$^b$: \#Q denotes the number of quadruplets.}\\
  \multicolumn{6}{l}{$^c$: denotes entity-aspect co-occurrence Pct. of sentences.} \\
  \multicolumn{6}{l}{$^d$: N Pct. denotes the Pct. of newly supplemented quads.}\\
\end{tabular}
 \vspace{-1mm}
\caption{  \label{tab:data_stat_coexist}
 \vspace{-1mm}
Statistics of newly constructed datasets.}
 \end{table}
% \end{wraptable}

\begin{table*}[htbp]
\vspace{-1mm}
  %\vspace{-3mm}\multirow{2}{*}{\bf }\thead{Method}
  \small
  \begin{tabular}{p{3.9cm}p{3.3cm}p{3.2cm}p{3.9cm}}
    \toprule
     \textbf{Examples}  & \textbf{ASTE} & \textbf{EASQE} & \textbf{Differential Attributions}   \\
    \midrule
    The staff offers impeccable service. &
    (\red{staff}, impeccable, NEG), 
    (\red{service}, impeccable, NEG)&
    (\blue{staff}, \blue{service}, impeccable, NEG) &
    \textbf{Entity-aspect co-occurrence}: biased generalization due to missing opinion extraction targets. \\
    \hline
    I love the easy to see screen , and It works ... &
    (\red{screen}, \red{easy}, POS)
    \red{or} 
    (\red{screen}, \red{easy to see}, POS)
    &
    (\blue{screen}, \blue{see}, easy, POS) &
    \textbf{Entity-aspect co-occurrence}: missing information or non-exclusive annotations. \\
    \hline
    Prices too high for this cramped and unappealing restaurant. &
    \red{-} &
    (\blue{restaurant, null, cramped, NEG} ), 
    (\blue{restaurant, null, unappealing, NEG}) & 
    \textbf{Implicit aspects}: missing extraction results containing implicit aspects. \\
    \hline
    The reflectiveness of the display is only a minor inconvenience if you ...
    % work in a controlled-lighting environment like me ( I prefer it dark ) or if you can crank up the brightness. 
    &
    (display, \red{minor inconvenience}, NEG) &
    (display, \blue{reflectiveness}, \blue{inconvenience}, NEG) &
    \textbf{Entity-aspect co-occurrence and Dataset quality}: inconsistent boundary extraction due to redundant degree adverbs.  \\
    \hline
    The PhotoBooth is a great program, it ... &
    (PhotoBooth, great, POS), 
    (\red{program, great, POS})  &
    (PhotoBooth, null,  great, POS) &
    \textbf{Dataset quality}: biased opinion extraction towards generalization. 
    \\
    \hline
    The cafe itself was really nice ...
    %with comfortable outdoor chairs and tables,
    but the service could have been better. &
    (service, \red{better}, NEG) &
    (null, service, \blue{could have been better}, NEG) &
    \textbf{Dataset quality}: missing negative expressions in opinion terms leading to mismatch with sentiment polarity. \\
    \hline
    %The pizza is delicious - 
    ... they use fresh mozzarella instead of the cheap, frozen, shredded cheese common to most pizzaria's.	 & 
    (\red{cheese, cheap, NEG }), 
    (\red{cheese, frozen, NEG }),
    (\red{cheese, shredded, NEG }) & 
    \blue{-} &
    \textbf{Dataset quality}: extracted opinions related to entities other than the reviewed object. \\
  \bottomrule
\end{tabular}
\vspace{-1mm}
  \caption{  \label{tab:cases}
  Examples comparing the difference of extracted opinions between our new EASQE dataset and the ASTE benchmark~\cite{wu-etal-2020-grid}. }
\vspace{-5mm}
\end{table*}

\section{Introduction}
{
Aspect-Based Sentiment Analysis (ABSA)~\cite{liu2012sentiment, pontiki-etal-2014-semeval} focuses on automatically extracting structured opinion related information from review text.
%, and is crucial for fine-grained sentiment analysis.
% including Aspect Sentiment Triplet Extraction (ASTE)~\cite{yan-etal-2021-unified, chen-etal-2022-enhanced} 
%Based on the combination of aspect terms, sentiment polarity, opinion terms, and aspect category 
% There exists six subtasks in ABSA. 
We summarize the subtasks of ABSA as follows: Aspect-level Sentiment Classification (ALSC), Aspect-oriented Opinion Extraction (AOE), Aspect Term Extraction and Sentiment Classification (AESC), (Aspect Term, Opinion Term) Opinion Pair extraction (OPE), (Aspect Term, Opinion Term, Sentiment Polarity) Opinion Triplet extraction (ASTE), and (Aspect Category, Aspect Term, Opinion Term, Sentiment Polarity) quadruplet prediction (ASQP)\footnote{It is also referred as Aspect-Category-Opinion-Sentiment (ACOS) Quadruple Extraction}. 
% • \cparagraph{ALSC}: Aspect-level Sentiment Classification.\\
% • \cparagraph{AOE}: Aspect-oriented Opinion Extraction.\\
% • \cparagraph{AESC}: Aspect Term Extraction and Sentiment Classification.\\
% • \cparagraph{OPE}: (Aspect Term, Opinion Term) Opinion Pair extraction.\\
% • \cparagraph{ASTE}: (Aspect Term, Opinion Term, Sentiment Polarity) Opinion Triplet extraction.\\
% • \cparagraph{ASQP}\footnote{It is also referred as Aspect-Category-Opinion-Sentiment (ACOS) Quadruple Extraction}: (Aspect Category, Aspect Term, Opinion Term, Sentiment Polarity) quadruplet prediction. \\
All the subtasks simplified the tasks by flattening entity-aspect nested relations~\cite{liu2012sentiment} and give a collective definition of the ``\emph{Aspect Term}'', but the methods are flawed in uniqueness and completeness.
% All the subtasks take the aspect term as their fundamental elements. 
% The aspect term given here is , thus introducing ambiguity in sentiment analysis. 
For the first example in Table~\ref{tab:cases}, ``\emph{The staff offers impeccable service}'', the aspect terms are ``\emph{staff}'' and ``\emph{service}'',
and the corresponding opinion term is ``\emph{impeccable}''. 
However, the aspect term ``\emph{service}'' is a specific attribute of ``\emph{staff}'' and does not represent everything else in the restaurant. 
Thus, the annotation of ASTE for \textbf{entity-aspect co-occurrence} cases such as  in the first row and second column in Table~\ref{tab:cases} may introduce bias towards generalization.
Moreover, the combined definition of aspect terms sometimes makes annotated labels not exclusive or information missing.
For the second example in Table~\ref{tab:cases},  ``\emph{I love the easy to see screen}'', the aspect term can be ``\emph{screen}'' with its opinion term ``\emph{easy}'' or the aspect term can be ``\emph{screen}'' with its opinion term ``\emph{easy to see}''. 
As shown in the fourth column of Table~\ref{tab:data_stat_coexist}, the co-occurrence rate ranges from 3.32\% to 14.16\% on the datasets. \\
\textbf{Implicit aspects} is another common scenario in the current datasets, and we have observed frequent issues with missing annotations as the third example in Table~\ref{tab:cases}.
Other data quality issues include inconsistent extraction boundaries, biased generalization, missing negative expressions in opinion terms, and extracted opinions related to entities other than the reviewed object. We provide examples illustrating these issues in the fourth to sevens rows in Table~\ref{tab:cases}. As evidenced by the fifth column of able~\ref{tab:data_stat_coexist}, the rate of newly supplemented quads on the datasets ranges from 5.93\% to 16.42\%.

% We found that there are many cases of entity-aspect co-occurrence in product reviews.
% Furthermore, many opinions with implicit aspect terms are also ignored in previous studies, as shown in the fifth column of Table~\ref{tab:data_stat_coexist} and the fourth example in Table~\ref{tab:cases},  the supplemented quads rate ranges from 5.93\% to 16.42\% on the datasets.
% Moreover, previous studies have also overlooked many opinions that contain implicit aspect terms, 
%To achieve solving a better sentiment analysis structure in solving the aforementioned two issues including entity-aspect co-existence, and non-exclusive extractions , we propose a new Entity-Aspect-Opinion-Sentiment Quadruple Extraction task (EASQE). Here, the opinion target decomposed and described in a structured manner with both entity and aspect levels included, which dramatically facilitate both extractions of opinions and later uses of the extracted opinion results.

In order to address the aforementioned issues of entity-aspect co-occurrence and implicit aspects and to achieve a more effective sentiment analysis structure, we propose a new task called Entity-Aspect-Opinion-Sentiment Quadruple Extraction (EASQE). In this task, the opinion target is decomposed and described in a structured manner, with both entity and aspect levels included. 
% This approach \red{significantly facilitates} the extraction of opinions and the later use of the extracted opinion results.
EASQE separately generates an entity term ``\emph{staff}'' and an aspect term ``\emph{service}'' for the positive opinion ``\emph{impeccable}'', and forms a comprehensive opinion quadruplet as in the first example of Table~\ref{tab:cases}}. In dealing with non-exclusive issues, EASQE extracts a unique quadruplet with the entity term ``\emph{screen}'', the aspect term ``\emph{see}'' and its positive opinion term ``\emph{easy}'' in the second example of Table~\ref{tab:cases}.
In addition, our EASQE use ``\emph{null}'' as in the third example of Table~\ref{tab:cases} to represent implicit aspects or entities.

In the past works, there is no public dataset found that could be used
% no public dataset to be applied 
to verify the performance of our proposed trigger-opinion on EASQE. Though existing labeled datasets, such as SemEval Challenges~\cite{pontiki-etal-2014-semeval, pontiki-etal-2015-semeval,pontiki-etal-2016-semeval} and ASTE datasets~\cite{wu-etal-2020-grid,peng2020knowing,xu-etal-2020-position} have provided concise and coherent texts for sentiment analysis tasks, they do not annotate entity and aspect terms separately, and some annotated labels have inconsistency or other data quality issues (See Table~\ref{tab:cases}).  
Hence, we build and release four new EASQE datasets by refining and aligning the datasets from ~\cite{wu-etal-2020-grid} with original SemEval Challenge datasets. Moreover, we develop a framework called Trigger-Opinion to learn from EASQE datasets and apply to the ASTE tasks, to justify the generalization of our Trigger-Opinion in handling other ABSA tasks. 

We summarize the contributions of this work as follows: 
\begin{itemize}  [parsep=0pt]
\item[--]  We define and investigate a new task Entity-Aspect-Opinion-Sentiment Quadruple Extraction (EASQE), which covers opinion target entities and their attributes, 
 and propose a novel framework named Trigger-Opinion to tackle EASQE.  
\item[--]  We have manually annotated four datasets for the EASQE task in the restaurant and laptop domains, which will serve as valuable resources for future research in this area.
\item[--] We prove that the EASQE task can fully exploit the sentiment information in the reviews and produce an exclusive data structure that can effectively perform fine-grained semantic analysis.
% Extensive empirical experiments demonstrate the feasibility of Trigger-Opinion in EASQE task. 
% More importantly, we prove that EASQE task can fully utilize the semantics information in reviews and brings out an exclusive semantic structure, that works effectively in fine-grained semantic analysis. 
% More importantly, we prove that the EASQE task can fully exploit the sentiment information in the reviews and produce an exclusive data structure that can effectively perform fine-grained semantic analysis.

\end{itemize}

\section{Related Work}
{
Traditional sentiment analysis has been extensively studied on sentence-level~\cite{ lan2019albert, raffel2020exploring} or document-level~\cite{lyu-etal-2020-improving,dou2017capturing}. Recently, significant research effort has been invested into Aspect-based-sentiment analysis (ABSA) ~\cite{hu2004mining,pontiki-etal-2014-semeval, pontiki-etal-2015-semeval,pontiki-etal-2016-semeval} and other fine-grained sentiment analysis tasks ~\cite{ramanathan2019twitter}.
The ABSA subtasks mainly include Aspect Term Extraction (ATE)~\cite{wei-etal-2020-dont,chen-qian-2020-enhancing}, 
Aspect-level Sentiment Classification (ALSC)~\cite{zhang-qian-2020-convolution,li-etal-2021-dual-graph},
Aspect Term Extraction and Sentiment Classification (AESC)~\cite{luo-etal-2019-doer,chen-qian-2020-relation},
Aspect-oriented Opinion Extraction(AOE)~\cite{wu2020latent,pouran-ben-veyseh-etal-2020-introducing},
Opinion Pair Extraction (OPE)~\cite{dai-song-2019-neural,zhao-etal-2020-spanmlt},
Aspect Sentiment Triplet Extraction (ASTE)~\cite{li2021sentiprompt,wu-etal-2020-grid,mao2021joint,yan-etal-2021-unified,xu-etal-2021-learning,zhang-etal-2022-structural,liu-etal-2022-robustly,chen-etal-2022-enhanced}
and Aspect-Category-Opinion-Sentiment Quadruple Extraction (ASQP)~\cite{cai-etal-2021-aspect,zhang-etal-2021-aspect-sentiment}.
% However, these tasks and methods extract entity and aspect terms in a combined form, which cannot distinguish opinion targets and their corresponding features in further analysis. 
Some works~\cite{wu-etal-2020-grid,xu-etal-2020-position,chen-etal-2022-enhanced,yan-etal-2021-unified} propose end-to-end models to jointly extract opinion triplets. 
However, the extracted aspect terms cannot be easily decomposed into entity and aspect terms. Other methods extract opinion pairs or triplets in pipelines~\cite{peng2020knowing}, but they conduct aspect extraction at the first stage, and the ambiguity of aspect terms even worsens the error propagation problem. 
Both of these methods do not distinguish between opinion targets and their corresponding attributes, which can limit further analysis.
This motivates us to explore sentiment analysis in a quaternary format including isolated entity terms.

Previously, nested relations of entity terms and aspect terms are simplified in sentiment analysis.
The main reason is that it is very difficult to identify parts and attributes of entities at different levels of detail~\cite{liu2012sentiment}.
Recently, many difficult NLP tasks such as event extraction~\cite{liu-etal-2020-event}, nested named entity recognition (NER)~\cite{li-etal-2020-unified}, and Spoken Language Understanding (SLU)~\cite{zeng2021automatic} can be effectively solved by sequence-tagging related methods due to the great progress of neural networks and pre-trained models.
This motivates us to investigate the nested relations of entity terms and aspect terms using sequence-tagging approaches for fine-grained sentiment analysis.
Moreover, Machine Reading Comprehension (MRC) models are integrated in solving ASTE tasks~\cite{mao2021joint, liu-etal-2022-robustly, chen2021bidirectional}, and effectively highlight the semantic relations between opinion terms and opinion targets.  
However, in their works, the question templates used in MRC are excessively long, particularly when compared to the short length of product reviews. 
Therefore, the effectiveness of attention models is obstructed when the sentence length increases.
Inspired by the aforementioned works, we design a new framework Trigger-Opinion to solve the new task with shorter trigger words instead of long question templates.
We believe that this work will open pathways for more intelligent opinion extractions and fine-grained sentiment analysis. 
}

\section{Problem Formulation}
\begin{figure*}
  \includegraphics[scale=0.26]{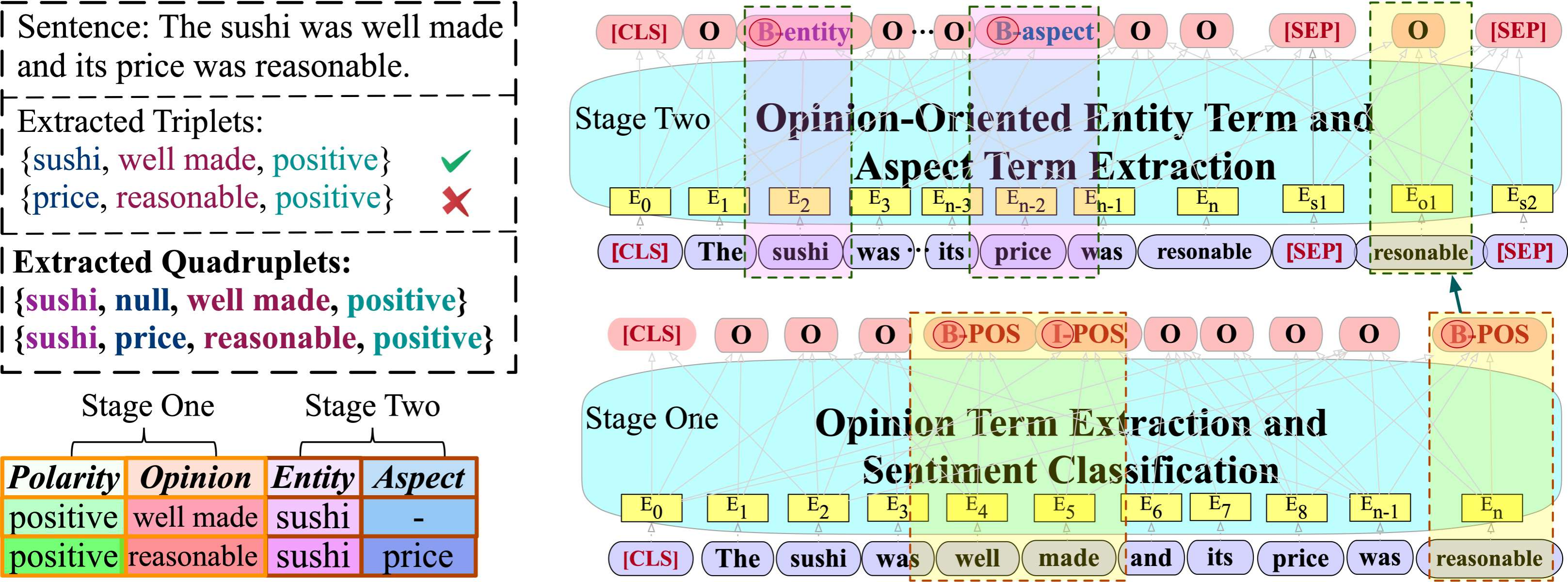}
  \centering
  %\vspace{-1mm}
  \caption{Proposed Trigger-Opinion framework.}
  %\vspace{-5mm}
  \label{fig:Trigger-Opinion}
\end{figure*}

\label{sec:task_formulation}
{
The task of Entity-Aspect-Opinion-Sentiment Quadruple Extraction is defined as follows: given a raw sentence,  $\s=w_1,w_2\ldots w_m$ denotes a sentence with $m$ tokens, our goal is to derive a set of 
entity-aspect-opinion-sentiment quadruplets $\CQ=\{(\e,\a,\o,\p)_{i}\}_{i=1}^Q$ from the sentence $s$, where 
$(\e,\a,\o,\p)_{i}$ is an opinion quadruplet in $s$.
The notations $e$ denotes an entity term,  $a$ denotes an aspect term, $o$ denotes an opinion term, and $p$ denotes the sentiment polarity and $\p\in\{\pos,\neu,\neg\}$.
An entity $e$ is a target object that has been evaluated, an aspect $a$ is an attribute or feature of the corresponding entity, an opinion $o$ is an evaluating expression for the corresponding entity and aspect ~\cite{liu2012sentiment}. 
$e$, $a$ and $o$
% These three terms 
are textual mentions extracted from sentences, and $e$ and $a$ can be ``\emph{null}''  if the target or attribute is not explicitly mentioned. 
}

\section{Proposed Framework}
{
To attain EASQE quadruplet element labels as defined in Section~\ref{sec:task_formulation}, we train a two-stage Trigger-Opinion framework on an corpus with $L$ annotated utterances.  
Nowadays, BERT~\cite{devlin2018bert} has demonstrated its superior performance on many downstream NLP tasks.  
% We, therefore, apply it to tackle the EASQE task on both stage of Trigger-Opinion.
Therefore, we apply it on both stages of Trigger-Opinion to tackle EASQE.
We present the two-stage framework in Figure~\ref{fig:Trigger-Opinion}, and provide a detailed explanation of the training process in Section~\ref{sec:stage1} and Section~\ref{sec:stage2}.

\subsection{Stage One}
\label{sec:stage1}
Given a sentence $\s$ with $m$ tokens, we first train a sequence-tagging model to output the opinion terms and corresponding sentiment polarity in a label sequence $\r=r_1,r_2\ldots r_{m}$. 
We apply the Beginning-Inside-Outside (BIO) schema~\cite{ramshaw1999text} at this stage. 
Hence, $r_i$ can be selected from one of the seven tags, $r_i\in{R_t}$,
${R_t}=\{B-\pos, I-\pos, B-\neu, I-\neu, B-\neg, I-\neg, O\}$.
 More specifically, given the sentence $\s$, we denote it by 
\[
[\CLS]\,w_1\,w_2\,\ldots\,w_{m}\,[\SEP],
\]
where $[\CLS]$ and $[\SEP]$ are two special tokens for the classification embedding and the separation embedding.  $w_i$ is the $i$-th subword extracted from BERT's dictionary.  After applying BERT's representation, 
we then apply a softmax classifier on top of the hidden features to compute the probability of each token $w_{i}$ to the corresponding label: %$r_{j}$  ($j=1, 2, 3, 4$):
\begin{equation}\label{eq:st}
p({r}_{i}|w_{i}) = \mbox{softmax}(\W \h_{i}),
\end{equation}
where $\W$ is the weight matrix.  
For CRF training, we use the maximum conditional likelihood estimation.
The logarithm of the likelihood is:
\begin{equation}\label{eq:crf}
\L(\W, b) = \sum_{i}\mbox{log}p({r}_{i}|w_{i};\W, b) 
\end{equation}
where $b$ is the bias.
\begin{equation}\label{eq:crf-max}
\r^* = \mathop{argmax}_{r\in{r_s}}p({r}_{i}|w_{i};\W, b) 
\end{equation}
Searching for the label sequence $\r$ with the highest conditional probability, opinion terms and sentiment polarity can be obtained by the BI-tags. This framework can also be utilized to solve other ABSA tasks. For OPE, ${r_s}=\{B, I, O\}$, and other parts remain the same; while ASTE has exactly the same framework in stage one as EASQE.

\subsection{Stage Two}
\label{sec:stage2}
We can obtain a list of opinions $\CO=(o_1 \cdots o_{n})$ in stage one. For each opinion term $\o_l\in{\CO}$, $\o_l = w_u\ldots w_{v}$, where $1 \leq u \leq v \leq m$, we concatenate it to the end of the original sentence as the trigger word, and train a sequence-tagging model to output the entity terms and aspect terms in a label sequence $\d=d_1,d_2\ldots d_{m}$. 
Similarly, using BIO scheme, $d_i$ can be selected from one of the five tags, $d_i\in{D_t}$,
${D_t}=\{B-\entity, I-\entity, B-\aspect, I-\aspect, O\}$.
 More specifically, given the sentence $\s$ and opinion term $\o_j$, we denote it by 
\[
[\CLS]\,w_1\,w_2\,\ldots\,w_{m}\,[\SEP]\,w_{u}\,\ldots\,w_{v}\,[\SEP].
\]
Using the same BERT representation, with the same training framework followed by the same softmax classifier and CRF algorithm in Section~\ref{sec:stage1}, we can obtain the entity terms and aspect terms by the BI-tags. 
This framework can also be utilized to solve other ABSA tasks. For OPE and ASTE, ${D_t}=\{B-\aspect, I-\aspect, O\}$, and other parts remains the same. 

\subsection{Quadruplet Decoding}
\label{sec:inference}

\begin{algorithm}[ht!]
\caption{Quadruplet Decoding for EASQE} 
\label{alg:Framwork} 
\begin{algorithmic}[1] 
\Require 
The input sentence $\s$; $\CE$ denotes the predicted entity terms, $\CA$ denotes the predicted aspect terms, $\CO$ denotes the predicted opinion terms, $\CP$ denotes the predicted sentiment polarity labels.$\CE=(e_1\cdots e_{q})$, $\CA=(a_1 \cdots a_{k})$, $\CO=(o_1 \cdots o_{n})$, $\CP=(p_1 \cdots p_{n})$. $GetTgt$ is the opinion targets inference function given the opinion terms.
% \State $\{(o_1, p_1) \cdots, (o_n, p_n)\}\leftarrow GetOpn(\s)$
% \State $\{(e_1\cdots e_{m}),(a_1 \cdots a_{k})\}\leftarrow GetTgt(\s, o_x)$
% \\\Return $\CQ$
\State $Tuple=\{\}$
\State $\CQ=\{\}$
\For{$i=1$; $i\leq{n}$; $i++$} 
\If{$GetTgt(\CP, \CO) \mbox{ is not empty}$}
\State $(\CE, \CA) = GetTgt(\CP, \CO)$
\If{$q\times k==1$}
\State
$Tuple = \{e_0, a_0, o_i, p_i\}$ 
\If{$Tuple \mbox{ not} \in \CQ$}
\State 
$\CQ=\CQ\cup Tuple$
\EndIf
\ElsIf{$k\geq1$}
\For{$j=1$; $j\leq{k}$; $j++$}
\State 
$Tuple = \{null, a_j, o_i, p_i\}$  
\If{$Tuple \mbox{ not} \in \CQ$}
\State 
$\CQ=\CQ\cup Tuple$
\EndIf
\EndFor
\ElsIf{$q\geq1$}
\For{$j=1$; $j\leq{q}$; $j++$}
\State 
$Tuple = \{e_j, null, o_i, p_i\}$  
\If{$Tuple \mbox{ not} \in \CQ$}
\State 
$\CQ=\CQ\cup Tuple$
\EndIf
\EndFor
\EndIf
\EndIf
\EndFor
\\\Return $\CQ$
\end{algorithmic} 
\end{algorithm}
% \vspace{-5mm}

In the following, we outline how to apply the trained Trigger-Opinion framework to conduct sentiment analysis. Algorithm~1 outlines the procedure of inferring the tagging-labels to quadruplets. As Line~3-5 shows, the opinion term along with its corresponding sentiment polarity are predicted in pairs. Each opinion is then combined with the original sentence and used as input for the second stage.
As shown in lines~6-9, for the extracted entity and aspect items, they can be considered as a pair only if they have exactly one entity and one aspect.
Otherwise, the extracted terms are separated into aspect terms (as in line~10-14) and entity terms (as in line~15-19) belonging to different quadruplets.

For the ASTE task, the decoding part is different from the EASQE on line~6-9 of Algorithm~1. Specifically, we only take entity term $e_0$ into account, and the opinion triplet $\{e_0, o_i, p_i\}$ is appended to the $\CQ$. Besides, we follow Algorithm~1, and append the original $Tuple$ without $null$ to $\CQ$.
}

\section{Experiments}
\label{sec:exp_all}
In this section, we conduct experiments on the new EASQE task and other ABSA subtasks to answer the following questions:
\begin{itemize}  [parsep=0pt]
\item[--]  \textbf{Q1:} Why Trigger-Opinion? How does our Trigger-Opinion perform on the EASQE task and other ABSA subtasks?
\item[--]  \textbf{Q2:} 
%What is the effect of decomposing opinion targets into entity terms and aspect terms, and to what extent does EASQE yield more comprehensive opinion extraction for fine-grained sentiment analysis? 
How effective is the decomposition of opinion targets into entity terms and aspect terms, and to what extent does EASQE provide more comprehensive sentiment analysis?
{\item[--]  \textbf{Q3:} Why new datasets and what is the quality of the new datasets for EASQE?}
\end{itemize}

\begin{table*}[htp]
\centering
   \small
  \begin{tabular}{c|c|c|c|c}
    \toprule
    \multirow{2}{*}{Methods} & {\bf Res14-EASQE} &  {\bf Lap14-EASQE} & {\bf Res15-EASQE} &{\bf Res16-EASQE} \\
    \cline{2-5} 
     & P/R/F1 & P/R/F1  & P/R/F1 & P/R/F1  \\
    \hline
    \multicolumn{5}{l}{\textbf{the OPE task}} \\
    \hline
    GTS$^\S$ & 74.56/74.57/74.55 & 67.49/60.21/63.57 & 70.09/64.24/66.89 & 74.81/74.56/74.68  \\
    BARTABSA$^\S$ & 72.31/72.01/72.15 & 61.37/55.59/58.30 & 68.54/\underline{65.80}/67.14 & 72.23/72.41/72.32  \\
    dual-MRC$^\natural$ & \underline{80.05}/\underline{76.77}/\underline{78.38} & \textbf{73.08}/\underline{64.52}/\underline{68.52} & \underline{72.46}/65.07/\underline{68.56} & \textbf{78.00}/\underline{77.23}/\underline{77.61}  \\
    Trigger-Opinion & \textbf{80.69}/\textbf{82.25}/\textbf{81.46} & \underline{71.34}/\textbf{72.06}/\textbf{71.69} & \textbf{73.58}/\textbf{74.42}/\textbf{73.98} & \underline{77.01}/\textbf{81.72}/\textbf{79.18}  \\
    \hline
    \multicolumn{5}{l}{\textbf{the ASTE task}} \\
    \hline
    GTS$^\S$ & 71.98/70.51/71.22 & 64.09/56.35/59.78 & 67.86/59.60/63.40 & 71.81/67.08/69.36  \\
    BARTABSA$^\S$ & 68.72/68.43/68.57 & 58.40/52.89/55.48 & 65.47/62.86/64.14  & 68.94/69.09/69.00  \\
    dual-MRC$^\natural$ & 76.63/72.58/74.54 & 65.66/59.26/62.29 & 68.78/59.75/63.97 & 71.65/70.55/71.10  \\
    Span-ASTE$^\S$ & \textbf{80.77}/73.36/76.88 & \underline{67.24}/62.54/64.75 & \underline{71.05}/67.06/68.97 & \textbf{76.85}/70.62/73.54  \\
    Trigger-Opinion-T & 77.28/\underline{79.13}/\underline{78.19} & \textbf{69.42}/\underline{66.49}/\underline{67.92} & 69.65/\underline{69.49}/\underline{69.56} & \underline{72.87}/\underline{75.62}/\underline{74.22}  \\
    Trigger-Opinion-QT & \underline{77.51}/\textbf{79.62}/\textbf{78.54} & 66.99/\textbf{69.74}/\textbf{68.33} & \textbf{71.65}/\textbf{70.29}/\textbf{70.96} & 72.07/\textbf{77.13}/\textbf{74.50}  \\
    \hline
    \multicolumn{5}{l}{\textbf{the EASQE task}} \\
    \hline
    dual-MRC$^\natural$ & \underline{73.43}/\underline{70.09}/\underline{71.72} & \underline{57.45}/\underline{51.97}/\underline{54.57} & \underline{63.59}/\underline{57.72}/\underline{60.51} & \underline{66.90}/\underline{65.96}/\underline{66.42}  \\
    Trigger-Opinion & \textbf{74.67}/\textbf{76.73}/\textbf{75.68} & \textbf{58.20}/\textbf{60.98}/\textbf{59.57} & \textbf{68.45}/\textbf{67.12}/\textbf{67.77} & \textbf{69.93}/\textbf{73.97}/\textbf{71.89}  \\
  \bottomrule
\end{tabular}
\vspace{-1mm}
 \caption{   \label{tab:new-res}
 Compared results on newly constructed EASQE datasets. We conduct experiments on three different tasks: OPE, ASTE, and EASQE. Best and second-best results are respectively in bold and underlined. 
 Trigger-Opinion-QT indicates that the framework is trained on the EASQE task and inferred on the ASTE task.
 Trigger-Opinion-T indicates the framework is both trained and inferred on the ASTE task. $^\S$ denotes that we reproduce the models using released code from the original papers.   $^\natural$ denotes that we reproduce the models taking reference to the original paper.}
 \vspace{-1mm}
\end{table*}

\subsection{Dataset}
The EASQE datasets (statistics are given in Table~\ref{tab:data_stat_coexist}) are built upon SemEval Challenges~\cite{pontiki-etal-2014-semeval, pontiki-etal-2015-semeval,pontiki-etal-2016-semeval} and ~\cite{wu-etal-2020-grid}. 
We align samples from these two sources and further conduct some annotations and refinements.
% \begin{itemize}  [parsep=0pt]
% \item[--] 

First, opinion targets including entity terms and aspect terms are combined as single aspect terms in ASTE tasks and datasets ~\cite{wu-etal-2020-grid}. Hence, we add entity terms to the original triplets for more complete semantic expressions. Take the sentence in Figure~\ref{fig:Trigger-Opinion} as an example, the original triplet \emph{(price, reasonable, positive)} does not include an entity term corresponding to the aspect term ``\emph{price}''. We annotate ``\emph{sushi}'' as the entity term and obtain the quadruplet \emph{(sushi, price, reasonable, positive)}.

Second, for other triples containing implicit entities or aspects, we manually annotate the original ``aspect'' as ``entity'' or ``aspect'', and leave the implicit element as ``null''.
% Secondly, for other triplets that have included implicit entity terms or aspects, we judge and manually annotate the original ``aspect'' into ``entity'' or ``aspect''. 
For example, the original triplet \emph{(sushi, well made, positive)} has an implicit aspect term. So we annotate ``\emph{sushi}'' as ``entity'', and obtain another quadruplet \emph{(sushi, null, well made, positive)}.

Finally, some annotated labels in ~\cite{wu-etal-2020-grid,peng2020knowing,xu-etal-2020-position} and other ASTE datasets have inconsistencies, missing annotations and other quality issues (See Table~\ref{tab:cases} and Appendix). 
In dealing with these issues, we define the following annotation instructions: 1) must include negative expressions in opinion terms; 2) exclude unnecessary degree adverbs in opinion terms; 3) exclude opinions related to entities other than the reviewed object. Other details about the annotation instructions will be released after publication. 
We refine the annotations from ~\cite{wu-etal-2020-grid} and construct four annotated datasets by three NLP experts with an inter-annotator agreement of the Fleiss’ Kappa equals to 0.77. 

To compare with current baselines in the ASTE tasks, the constructed EASQE datasets are then transformed back to triplet format, using the decoding algorithm as in Section~\ref{sec:inference}. 
% We further transform the EASQE datasets for the ASTE task, using the decoding algorithm as in Section~\ref{sec:inference}. 
Additionally, we remove sentiment polarities from opinion triplets, and obtain the datasets for the OPE task.
% \end{itemize}

\subsection{Baselines}
\label{sec:baselines}
Our Trigger-Opinion approach is compared to the following baselines:  \\
% Our Trigger-Opinion approaches are compared with the following baselines using pipeline or end-to-end methods: \\
\textbf{GTS:} ~\cite{wu-etal-2020-grid} addresses the ASTE and the OPE tasks in an end-to-end fashion with a unified grid tagging task. \\
\textbf{BARTABSA:} ~\cite{yan-etal-2021-unified} exploits the pre-trained sequence-to-sequence model BART to solve all ABSA subtasks.  \\
\textbf{Span-ASTE:} ~\cite{xu-etal-2021-learning} proposes a span-level approach to consider the interaction between the spans of targets and opinions, and a pruning strategy by incorporating supervision from the ASTE and the OPE tasks. \\
\textbf{dual-MRC:} ~\cite{mao2021joint} constructs two machine reading comprehension (MRC) tasks to solve OPE and ASTE subtasks. 
% \red{check}
We further edit the pattern question in the first MRC stage to apply this method to the EASQE task (see more details in Appendix).
% We edit the pattern questions at the target extraction MRC and add quadruplet inference rules to apply this method to EASQE task .

\subsection{Experiment Setup}
In training the Trigger-Opinion framework, we fine-tune the uncased BERT-base\footnote{\url{https://github.com/google-research/bert}} on train datasets by the following settings: a mini-batch size of 4, a maximum sequence length of 64, and a learning rate of $2 \times 10^{-5}$. 
The development datasets are used for early stopping. 
We run each model five times and report the average result of them.
All experiments are conducted on a single Tesla-V100-32G GPU.

For all tasks in our experiments, we use precision (P), recall (R), and F1-score as the evaluation metrics. An extracted term is considered correct if it exactly matches a golden term, and an extracted pair/triplet/quadruplet is considered correct if all elements of it exactly match the golden pair/triplet/quadruplet. 

\begin{table*}[htbp]
\centering
   \small
  \begin{tabular}{c|c|c|c|c}
    \toprule
    \multirow{2}{*}{Methods} & {\bf Res14} &  {\bf Lap14} & {\bf Res15} &{\bf Res16} \\
    \cline{2-5} 
     & P/R/F1 & P/R/F1  & P/R/F1 & P/R/F1  \\
    \hline
    Peng+IOG~\cite{wu-etal-2020-grid} & 58.89/60.41/59.64 & 48.62/45.52/47.02 & 51.70/46.04/48.71 & 59.25/58.09/58.67  \\
    IMN+IOG~\cite{wu-etal-2020-grid} & 59.57/63.88/61.65 & 49.21/46.23/47.68 & 55.24/52.33/53.75 & -  \\
    GTS-CNN~\cite{wu-etal-2020-grid} & 70.79/61.71/65.94 & 55.93/47.52/51.38 & 60.09/53.57/56.64 & 62.63/66.98/64.73 \\
    GTS-BiLSTM~\cite{wu-etal-2020-grid} & 67.28/61.91/64.49 & 59.42/45.13/51.30 & 63.26/50.71/56.29 & 66.07/65.05/65.56 \\ 
    $\rm S^{3}E^{2}$~\cite{chen-etal-2021-semantic} & 69.08/64.55/66.74 & 59.43/46.23/52.01 & 61.06/56.44/58.66 & 71.08/63.13/66.87  \\ 
    GTS-BERT~\cite{wu-etal-2020-grid} & \underline{70.92}/69.49/70.20 & 57.52/51.92/54.58 & \underline{59.29}/58.07/58.67 & \textbf{68.58}/66.60/67.58 \\
    BMRC ~\cite{chen2021bidirectional} & -/-/70.01 & -/-/57.83  & -/-/58.74 & -/-/67.49  \\
    EMC-GCN~\cite{chen-etal-2022-enhanced} & \textbf{71.85}/\underline{72.12}/\underline{71.98} & \textbf{61.46}/\underline{55.56}/\underline{58.32} & \textbf{59.89}/\underline{61.05}/\underline{60.38} & \underline{65.08}/\underline{71.66}/\underline{68.18} \\
    Trigger-Opinion & 70.19/\textbf{74.13}/\textbf{72.10} & \underline{60.40}/\textbf{58.82}/\textbf{59.59} & 58.72/\textbf{64.50}/\textbf{61.47} & 64.51/\textbf{74.01}/\textbf{68.93}  \\
  \bottomrule
\end{tabular}
\vspace{-1mm}
 \caption{   \label{tab:old-res}
 Compared results on ASTE datasets~\cite{wu-etal-2020-grid}. Best and second-best results are respectively in bold and underlined. All baseline results are from the original papers. }
 \vspace{-1mm}
\end{table*}

\subsection{Performance on the EASQE task}
Table~\ref{tab:new-res} reports the performance of the Trigger-Opinion framework on three subtasks, and it is evaluated on all four EASQE datasets to answer \textbf{Q1-2}. For comparison, we tune dual-MRC as in Section~\ref{sec:baselines} on the EASQE task as a baseline. We have the following observations:
\begin{itemize}  [parsep=0pt]
\item[--]  Trigger-Opinion significantly outperforms the baseline dual-MRC on EASQE task of all four datasets under the paired $t$-test ($p<0.05$). It attains an average of \textbf{5.40\%} higher F1-score than the baseline. 
\item[--]  By analyzing the results on the ASTE task, Trigger-Opinion achieves competitive performance when trained on EASQE datasets comparing with ASTE datasets. Specifically, Trigger-Opinion-QT attains an average of \textbf{0.61\%} higher F1-score than Trigger-Opinion-T using the same pretrained model and similar framework. 
This small improvement makes sense since EASQE is a relatively difficult task. 
The results demonstrate the necessity of decomposing opinion targets into entity terms and aspect terms in handling sentiment analysis.  
\item[--] 
% \red{check} 
% With an extra element ``entity'', EASQE yields more comprehensive opinion extraction structure than ASTE.
As described in the triplet decoding process in Section~\ref{sec:inference}, the merging of ``aspect terms'' and ``entity terms'' in quadruplet into single “aspect terms” in triplet can lose some information.            
%be seen as a form of dimensionality reduction. Because it is true that some information may be lost in this process.
% We can infer from the triplet decoding process in Section~\ref{sec:inference} that, the decoding process from quadruplets to triplets is a dimensionality reduction process.
% This implies that, quadruplet results obtained from EASQE can be easily transferred to ASTE triplets which would otherwise be difficult. 
Combined with the excellent Trigger-Opinion-QT results, we can conclude that, quadruple results obtained from EASQE can be easily mapped to ASTE triplets, but not vice versa.
% however it does not work well in the opposite way.
This again verifies the necessity of the entity-included quadruple extraction for a more comprehensive opinion structure.
\end{itemize}

\subsection{Performance on the ABSA subtasks}
As previously mentioned, the Trigger-Opinion framework can also be utilized to solve the ABSA subtasks. Table~\ref{tab:new-res} and Table~\ref{tab:old-res} report the performance of the Trigger-Opinion framework on the OPE and ASTE tasks, and evaluations from two different dataset resources to further answer \textbf{Q1}. We have the following observations:
\begin{itemize}  [parsep=0pt]
\item[--]  
% Trigger-Opinion significantly outperforms the baselines, including end-to-end and pipeline methods, on all four EASQE datasets for both OPE and ASTE tasks. 
Trigger-Opinion significantly outperforms baselines on all four EASQE datasets for OPE and ASTE tasks, including end-to-end and pipeline methods.
% Another finding is that Trigger-Opinion has achieve more significant improvements on recall, which is an average of at least \textbf{6.72\%} and \textbf{5.80\%} higher on OPE and ASTE, respectively. 
Another finding is that Trigger-Opinion achieves more significant improvements in recall, averaging at least \textbf{6.72\%} and \textbf{5.80\%} higher on the OPE and the ASTE tasks, respectively.
% This improvement is attributed to that our Trigger-Opinion can leverage the relations between opinion terms and opinion targets. 
This improvement is attributed to the fact that our Trigger-Opinion can exploit the relations between opinion terms and opinion targets.
% More significantly, Trigger-Opinion can easily transfer to other ABSA subtasks such as OPE ans ASTE, and achieves the best performance.   
In conclusion, Trigger-Opinion can be easily transferred to other ABSA subtasks like OPE and ASTE and achieves the best performance.
\item[--]  Trigger-Opinion achieves more competitive performance than dual-MRC with similar pipeline structures. 
The results demonstrate: 1) the effectiveness of handling opinion terms extraction and sentiment classification in the first stage, 2) the significance of trigger words. 
Additional details of frame analysis are presented in Section~\ref{sec:analysis}.
\item[--]  By examining the results of previous ASTE benchmark~\cite{wu-etal-2020-grid}, Trigger-Opinion attains an average of \textbf{0.81\%} higher F1-score on ASTE than the best baseline. 
%Furthermore, Trigger-Opinion is a simpler method with better interpretability and lower resource requirements. 
\end{itemize}

\subsection{Framework Analysis}
\label{sec:analysis}
In the following, we conduct experiments on the current ASTE benchmark~\cite{wu-etal-2020-grid}, analyze the effect of each module of Trigger-Opinion, and further answer \textbf{Q1}.

\begin{figure}[htbp]
  \includegraphics[scale=0.21]{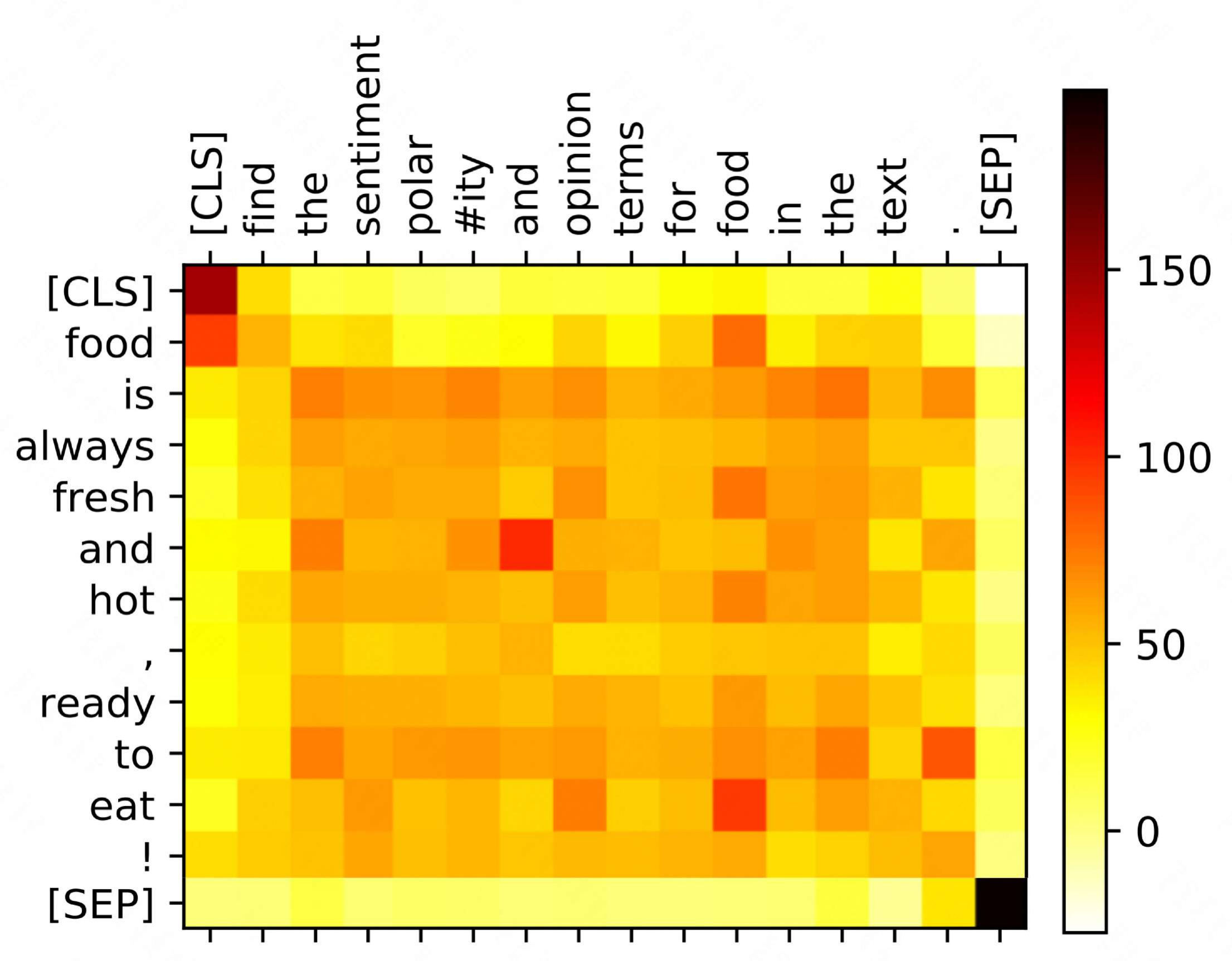}
  \centering
  \caption{
  %Visualization of attention matrices of an MRC pattern question and a product review.
  Attention matrix visualization for an example MRC pattern questions and a product review.}
  \label{fig:attention}
\end{figure}

\cparagraph{Effect of Trigger Words}
We plot in Figure~\ref{fig:attention} to visualize the attention matrix for an example MRC pattern question and a product review. 
It can be observed that the opinion target term ``\emph{food}'' has high attention scores with opinion terms ``\emph{fresh}'', ``\emph{hot}'' and ``\emph{ready to eat}''. 
Other words such as ``opinion terms'' and ``sentiment polarity'' in MRC question patterns also have relatively high attention scores with opinion terms, but not as obvious as opinion target words.
% This implies that the target terms are effective enough as trigger words. 
This implies that the target word is sufficiently effective as a trigger word.
Inspired by the idea of utilizing attention matrices to induce the relations between opinion terms and targets, and driven by the desire to minimize the sequence length and maximize the attention impact, we propose the new Trigger-Opinion framework. 

\cparagraph{Ablation Analysis} 
% In the following, we conduct experiments on the ASTE benchmark~\cite{wu-etal-2020-grid} to analyze the effect of each module in Trigger-Opinion, and justify its scheme designs.
% In the following, we conduct the OPE task experiments on the ASTE benchmark~\cite{wu-etal-2020-grid} to demonstrate the effectiveness of scheme design in Trigger-Opinion.
% to analyze the effects of each module in Trigger-Opinion and demonstrate the effectiveness of its scheme design.
% The experimental results are shown in Table~\ref{tab:analysis} denotes the Trigger-Opinion uses different extraction order in two stages and different trigger pattern. 
In the following, we conduct experiments and ablation analysis on Trigger-Opinion with different extraction orders and different trigger modes, and the results are shown in Table~\ref{tab:analysis}.
% The experimental results indicating that Trigger-Opinion uses different extraction orders and different trigger modes in the two stages are shown in Table 4.
Here, we implement a partMRC model that concatenates  ``aspect'' or ``opinion'' + trigger word to the end of product reviews.
% We conduct experiments on partMRC to investigate into the effect of other words in MRC pattern question in comparison with Trigger Words, and verify our observations in Figure~\ref{fig:attention}.  
We conduct experiments on partMRC to investigate the effect of the second most important words of MRC pattern questions except our trigger words, and to validate our observations in Figure ~\ref{fig:attention}.
Other experiments are designed for different extraction orders and sequence tagging scheme choices. 
The observations of Table~\ref{tab:analysis} are as follows:
\begin{itemize}  [parsep=0pt]
\item[--]  PartMRC performs well on the OPE task, but its F1 score is still \textbf{1.35\%} lower on average compared to Trigger-Word methods.
% still has an average of 2.21\% lower F1 compared with Trigger- Opinion.
% This implies that the extracted terms themselves are effective enough for the purpose of relationship utilizing in opinion extraction tasks. 
This shows that the extracted terms themselves are effective enough for the purpose of utilizing the target-opinion relations in the sentiment analysis task.
\item[--]  The span-based Trigger-Opinion framework has an average of \textbf{1.81\%} lower F1 compared with BIO-based Trigger-Opinion. 
This implies that BIO scheme is more flexible and more suitable for sequence labeling and inference for this task. 
\item[--]  
% By comparing the results with aspect term extracted first or opinion term extracted first, we can conclude that opinion-first methods have 1.44\% to 1.73\% higher F1 scores. 
By comparing the results of the aspect extracted first and the opinion extracted first, we can conclude that the F1 score of the opinion-first methods have \textbf{1.58\%} higher F1 score in average.
% The results make sense because most of opinion terms are adjective words or phrases, and can be easier extracted with more obvious semantic and grammatical features. 
% The results make sense because most opinion terms are adjectives, which have more obvious semantic and grammatical features and are easier to extract.
The results make sense since most opinion words are adjectives with more obvious semantic and grammatical features and thus easier to extract.
% Handling a relatively easier problem at the first stage can effectively alleviate the potential error propagation in the pipeline solution.
Dealing with a relatively easy problem in the first stage can effectively mitigate potential error propagation in the pipeline solution.
\end{itemize}

\begin{table}[htbp]
   \small
  \begin{tabular}{c|c|c|c|c}
    \toprule
    \multirow{2}{*}{Methods} & {\bf 14Res} &  {\bf 14Lap} & {\bf 15Res} &{\bf 16Res} \\
    \cline{2-5} 
     & F1 & F1  & F1 & F1  \\
    \hline
    partMRC-a$^a$ & 73.59 & 67.63 & 66.67 & 74.54 \\
    partMRC-o$^b$ & 74.29 & \underline{68.47} & \underline{69.32} & 76.09 \\
    span-based$^c$ & \underline{75.59} & 68.15  & 68.60 & 74.62  \\
    Trigger-Aspect$^d$ & 74.36 & 66.61 & 69.34 & \underline{76.94} \\
    Trigger-Opinion & \textbf{76.97} & \textbf{69.06} & \textbf{70.67} & \textbf{77.48}  \\
  \bottomrule
  \multicolumn{5}{l}{$^a$: partMRC with aspect extracted at the first stage.}\\
  \multicolumn{5}{l}{$^b$: partMRC with opinion extracted at the first stage.}\\
  \multicolumn{5}{l}{$^c$: span-based Trigger-Opinion.} \\
  \multicolumn{5}{l}{$^d$: BIO-based model with aspect extracted at the first stage. }\\
\end{tabular}
 \caption{   \label{tab:analysis}
 Ablation analysis results of Trigger-Opinion on OPE task datasets~\cite{wu-etal-2020-grid}. Best and second-best results are respectively in bold and underlined. }
\end{table}

\subsection{Case Study}
\label{sec:case study}
% To demonstrate the necessity and superiority of the EASQE task and its datasets and to summarize our answer for \textbf{Q3}, we summarize some difference between two datasets in Table~\ref{tab:cases}. 
To demonstrate the necessity and superiority of the EASQE dataset and answer \textbf{Q3}, we analyze the differences between EASQE and ASTE datasets in Table ~\ref{tab:cases}.
From the presented cases, we can see that the ASTE task lacks the ability to : 1) decompose the opinion target into entities and aspects, see the first, second, and fourth example;
2) conduct opinion extractions when the target is the default product or shop and has implicit aspect, see the third example; 
% When the target is the default product or store and has implicit aspects, perform opinion extraction, see Example 4;
3) distinguish the opinion target from its comparison object, see the seventh example.
% To remedy the above deficiencies and correct the other errors in Table~\ref{tab:cases}, we calibrate the new EASQE datasets for further studies.
To remedy the above deficiencies and to correct other errors in Table ~\ref{tab:cases}, we calibrated the new EASQE dataset for further study.

\section{Conclusions and Future Work}
In this paper, we define a new task of Entity-Aspect-Opinion-Sentiment Quadruple Extraction (EASQE) for more comprehensive sentiment analysis.  
% We then propose an simple yet effective approach, Trigger-Opinion, which can exploit relations between opinion terms and opinion targets.  
Then, we propose a simple and effective framework, Trigger-Opinion, which exploits the relations between opinion terms and opinion targets.
% More importantly, our Trigger-Opinion can be effectively transferred to other subtasks and achieves satisfactory results.
% Extensive experiments validate the feasibility and power of our model, and set a benchmark performance for the newly proposed EASQE task.
Extensive experiments validate the feasibility and robustness of our model and set a baseline performance for the newly proposed EASQE task. 
Moreover, our Trigger-Opinion can be effectively modified to solve other ABSA subtasks with satisfactory results. 
This work sheds light on the development of more comprehensive fine-grained sentiment analysis.  
Several promising future directions can be considered: (1) detecting 
sentiment elements including opinion target category; 
% (2) utilizing external well-defined knowledge graph or other methods to cluster the extracted opinion quadruplets; 
(2) clustering the extracted opinion quadruples using externally well-defined knowledge graphs or other resources;
(3) developing a generative formulation for sentiment analysis based on Trigger-Opinion.

% \clearpage

\section{Limitations}
% Our proposed method can not directly pair multiple (more than one) entity items and multiple (more than one) aspect items of the same opinion item into distinct quadruplets. 
Multiple entity terms and multiple aspect terms of the same opinion term can not be directly matched using our proposed method.
% Our proposed method does not directly match multiple entity terms and multiple aspect terms of the same opinion term into different quadruplets.

\section{Ethics Statement}
For the consideration of ethical concerns, we make detailed description as follows: \\
(1) All existing dataset sources are from public scientific papers.  \\
(2) The newly constructed datasets do not include any protected information.   \\
(3) We describe the statistics of the datasets in specific sections. Our analysis and results are consistent.    \\
(4) Our work does not contain identity characteristics.  \\
(5) Our experiments do not need a lot of computer resources compared to pretrained models. \\

\section*{Acknowledgements}

% Entries for the entire Anthology, followed by custom entries

\bibliography{acl_latex}

\clearpage
\appendix

\section{Appendix}
\begin{figure}[htbp]
  \includegraphics[scale=0.33]{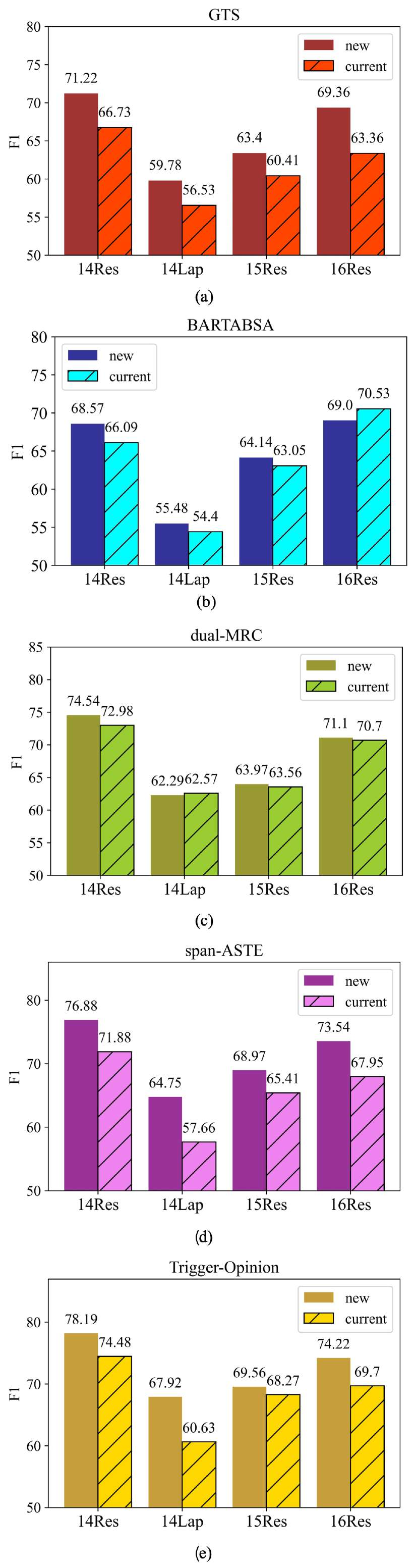}
  \centering
  \caption{Performance comparison between different training source using (a) GTS, (b) BARTABSA, (c) dual-MRC, (d) span-ASTE and (e) Trigger-Opinion methods.}
  \label{fig:append}
\end{figure}

\subsection{Modified Dual-MRC for EASQE}
\label{sec:MRC for quads}
We modify the first pattern question of dual-MRC as ``\emph{find the entity and aspect terms in the text}'', and keep the second pattern question unchanged.
Since both entity terms and aspect terms are extracted in the first stage, we enumerate each term and add the term to the second pattern question. 

\subsection{Quality Analysis on two datasets}
\label{sec:dataset quality}

Three NLP experts annotate different triplets between EASQE datasets and the previous ASTE dataset ~\cite{wu-etal-2020-grid}, and make a conclusion that EASQE contains more coherent results.
The Inter-annotator agreement is high (Fleiss’ Kappa larger than 0.81). 
% Moreover, we conduct comparison experiments between two data sources, with the same dev and test datasets from new EASQE, and train from different dataset respectively. 
Furthermore, we conduct comparative experiments between two data sources, using the same development and test datasets from the new EASQE and training from different datasets respectively.
Five different methods (GTS, BARTABSA, dual-MRC, span-ASTE and Trigger-Opinion) are included as in Figure~\ref{fig:append}. 
% In conclusion, only dual-MRC 14Lap and BARTABSA 16Res has better F1 score trained on ASTE dataset, other cases 
For the five different methods, the EASQE training dataset achieves an average of \textbf{4.18\%}, \textbf{0.78\%}, \textbf{0.52\%}, \textbf{5.31\%}, and \textbf{3.90\%}  higher F1 scores on the ASTE task, respectively.
%EASQE datasets attains an average of 4.18\%, 0.78\%, 0.52\%, 5.31\%, and 3.90\% higher F1-score on the ASTE task for the five different methods respectively. 
% This implies that EASQE datasets has higher quality in comparison with the previous ASTE datasets.  
This implies that EASQE dataset is of higher quality compared to the previous ASTE dataset.

\subsection{Dataset Release}
The EASQE datasets as well as the annotation guidelines and Trigger-Opinion code will be released after publication.

\end{document}